\documentclass[a4paper]{article}

\usepackage{INTERSPEECH2019}

\title{AN ONLINE ATTENTION-BASED MODEL FOR SPEECH RECOGNITION}
\name{Ruchao Fan$^{1,3}$, Pan Zhou$^{2}$, Wei Chen$^{3}$, Jia Jia$^{2}$, Gang Liu$^{1}$}
\address{
  $^1$ Beijing University of Posts and Telecommunications, Beijing, P.R.China\\
  $^2$Department of Computer Science and Technology, Tsinghua University, Beijing, P.R.China\\
  $^{3}$Voice Interaction Technology Center, Sogou Inc., Beijing, R.R.China}
\email{\{fanruchao,liugang\}@bupt.edu.cn, \{zh-pan,jjia\}@mail.tsinghua.edu.cn, chenwei@sogou-inc.com}

\begin{document}

\maketitle
\begin{abstract}
Attention-based end-to-end models such as Listen, Attend and Spell (LAS), simplify the whole pipeline of traditional automatic speech recognition (ASR) systems and become popular in the field of speech recognition. In previous work, researchers have shown that such architectures can acquire comparable results to state-of-the-art ASR systems, especially when using a bidirectional encoder and global soft attention (GSA) mechanism. However, bidirectional encoder and GSA are two obstacles for real-time speech recognition. In this work, we aim to stream LAS baseline by removing the above two obstacles. On the encoder side, we use a latency-controlled (LC) bidirectional structure to reduce the delay of forward computation. Meanwhile, an adaptive monotonic chunk-wise attention (AMoChA) mechanism is proposed to replace GSA for the calculation of attention weight distribution.  Furthermore,  we propose two methods to alleviate the huge performance degradation when combining LC and AMoChA. Finally, we successfully acquire an online LAS model, LC-AMoChA, which has only 3.5\% relative performance reduction to LAS baseline on our internal Mandarin corpus.
\end{abstract}

\section{Introduction}
\label{sec:intro}
Recently, end-to-end (E2E) models  have become increasingly popular on automatic speech recognition (ASR), as they allow one neural network to jointly learn acoustic, pronunciation and language model, greatly simplifying the whole pipeline of conventional hybrid ASR systems. Connectionist Temporal Classification (CTC) \cite{graves2006connectionist,amodei2016deep}, Recurrent Neural Network Transducer (RNN-T) \cite{graves2012sequence,battenberg2017exploring,rao2017exploring}, Recurrent Neural Aligner \cite{sak2017recurrent,dong2018extending}, Segment Neural Transduction \cite{yu2016online} and Attention-based E2E (A-E2E) models \cite{chorowski2014end, chan2016listen, chiu2018state} are such E2E models that are well explored in the literature. With a unidirectional encoder, these model architectures are easy to deploy for an online ASR task except the attention based model. 
A-E2E model was first introduced to ASR in \cite{chorowski2014end}. Later on, a paradigm named Listen, Attend and Spell (LAS) with a bidirectional long short-term memory (BLSTM) \cite{schuster1997bidirectional} encoder was examined on a large-scale speech task \cite{chan2016listen} and more recently it shows superior performance to a conventional hybrid system \cite{chiu2018state}. Besides, previous work showed that LAS offers improvements over CTC and RNN-T models \cite{prabhavalkar2017comparison}. However, LAS with bidirectional encoder and global soft attention (GSA) must process the entire input sequence before producing an output. This makes it hard to be used in online ASR scenarios. The goal of this paper is to endow the LAS with the ability to decode in a streaming manner with as less degradation of its performance as possible.

The first problem of LAS to be solved for online purpose is the computation delay in bidirectional encoder. Latency-controlled BLSTM (LC-BLSTM) was proposed \cite{zhang2016highway} to get a trade-off between performance and delay. This structure looks ahead for a fixed number of frames other than making use of  the entire future frames in BLSTM. With lower latency, LC-BLSTM hybrid system in \cite{xue2017improving} achieves comparable performance to BLSTM. However, it is not clear whether such structures would be practical for A-E2E models.

For GSA mechanism, the context vector for decoder is computed based on the entire encoder outputs. Thus, the model cannot produce the first character until all input speech frames have been consumed during inference. One straightforward idea for reducing the attention delay is to make the input sequence to be attended shorter. Neural Transducer (NT) \cite{jaitly2016online, sainath2018improving} is such a method. It cuts the input sequence into none-overlapping blocks and performs GSA over each block to predict the targets correspond to that block. The ground truth of each block needs to be acquired by a process named alignment, which resembles force alignment in conventional ASR systems and takes much time during training.
Another idea is to take advantage of the monotonicity of speech. Hard monotonic attention which attends only one of the encoder outputs was introduced to stream the attention part. The hard attention needs to calculate a probability for each encoder output to determine whether it should be attended or not at each step and this causes a non-differentiable problem. One solution for this problem is reinforcement learning \cite{luo2017learning,lawson2018learning} while the other is to estimate the expectation of context vector during training \cite{raffel2017online}. To better exploit the information of encoder, Chiu et al. proposed Monotonic Chunkwise Attention (MoChA) \cite{chiu2017monotonic} which adaptively splits the encoder outputs into small chunks over which soft attention is applied and only one output is generated for each chunk. However, MoChA uses a fixed chunk size for all output steps and all utterances. It is not appropriate for ASR due to the variational pronunciation rate of output units and speakers. 

In this work, we construct a LAS baseline system with BLSTM encoder and GSA mechanism, denoted as BLSTM-GSA, and explore methods to stream it by overcoming the abovementioned problems. On the encoder side, we replace BLSTM with LC-BLSTM to reduce the delay of forward computation and name the model as LC-GSA. With initialization from BLSTM-GSA, LC-GSA has almost no degradation to LAS baseline. On the attention side, MoChA with BLSTM encoder (BLSTM-MoChA) is explored and it also behaves well. 
Meanwhile, we propose an adaptive monotonic chunk-wise attention (AMoChA) to adaptively generate chunk length to better fit the speech properties. The AMoChA is different from the MAtChA in \cite{chiu2017monotonic}, which calculate expectation of weight distribution of encoder output to implement adaptive chunk length. On the contrary, we use several feed forward layers to predict the chunk length directly. Furthermore, we offer another two revisions which can be used both on MoChA and AMoChA to alleviate the degradation caused by combing the LC-BLSTM encoder with AMoChA (demoted as LC-AMoChA). Finally, the model LC-AMoChA can decode in a streaming manner with acceptable performance reduction of relative 3.5\% to offline model BLSTM-GSA.

The remainder of the paper is structured as follows. Section \ref{sec:basiclas} introduces the basic offline LAS model. The details of our methods to stream LAS baseline are described in Section \ref{sec:onlinelas}. Experimental details and results are presented in Section \ref{sec:results} and the paper is concluded with our findings and future work in Section \ref{sec:conclusion}.

\section{BASIC LAS MODEL}
\label{sec:basiclas}

The basic LAS model consists of three modules and it can be described by Eq.(\ref{eq:listener}) - (\ref{eq:outputPost}). Given a sequence of input frames $\boldsymbol{x}=\{x_1,x_2,\dots,x_T\}$, a listener encoder module, which consists of a multi-layer BLSTM is used to extract higher level representations $\textbf{h}=\{h_1,h_2,\dots,h_U\}$ with $U \leq T$.
\begin{equation}
\boldsymbol{h}=Listener(\boldsymbol{x})
\label{eq:listener}
\end{equation}
\begin{equation}
s_i = SpellerRNN(s_{i-1},y_{i-1},c_{i-1})
\end{equation}
\begin{equation}
e_{i,u}=Energy(s_{i},h_u) = V^T tanh(W_h h_u+W_s s_i+b)
\label{eq:energy}
\end{equation}
\begin{equation}
\alpha_{i,u} = \frac{exp(e_{i,u})}{\sum_{u'} exp(e_{i, u'})}
\label{eq:wts}
\end{equation}
\begin{equation}
c_i =\sum_{u}\alpha_{i,u}h_u
\label{eq:context}
\end{equation}
\begin{equation}
P(y_i|\boldsymbol{x},y_{<i}) = SpellerOut(s_i,c_i)
\label{eq:outputPost}
\end{equation}

The speller usually contains an LSTM layer and an output softmax layer. At every output step $i$, the attender first computes the context vector $c_i$ based on the speller state $s_i$ and listener output $\boldsymbol{h}$, and then the speller generates output distribution. Eq.(\ref{eq:energy}) - (\ref{eq:context}) tells how the attender works, where $W_h$, $W_s$, $V$ and $b$ are weights to be learned. This is also known as soft attention and visualized in Fig.\ref{fig:1}a. It tells the speller where to focus on in order to output the next unit.

\section{STREAMING METHODS}
\label{sec:onlinelas}

\subsection{Latency-controlled Listener}
\label{ssec:lcblstm}
LSTM is a straightforward way to stream the listener, but the performance gap is wide compared to BLSTM. 
Motivated by \cite{zhang2016highway,xue2017improving}, we attempt to use LC-BLSTM to replace BLSTM in the listener and try to make the performance loss as small as possible. 

LC-BLSTM has a better performance than unidirectional LSTM and a lower latency than BLSTM which can be tolerated by the online tasks in traditional hybrid systems. The main difference between LC-BLSTM and BLSTM is their data arrangement. For LC-BLSTM in our system, a sentence is first split into non-overlapping blocks of fixed length $N_c$, then $N_r$ future frames are appended as right context for the reversed LSTM to obtain limited future information, which causes some loss compared to the full sequence data arrangement for BLSTM.

\begin{figure}[t]
\centerline{\includegraphics[width=\linewidth]{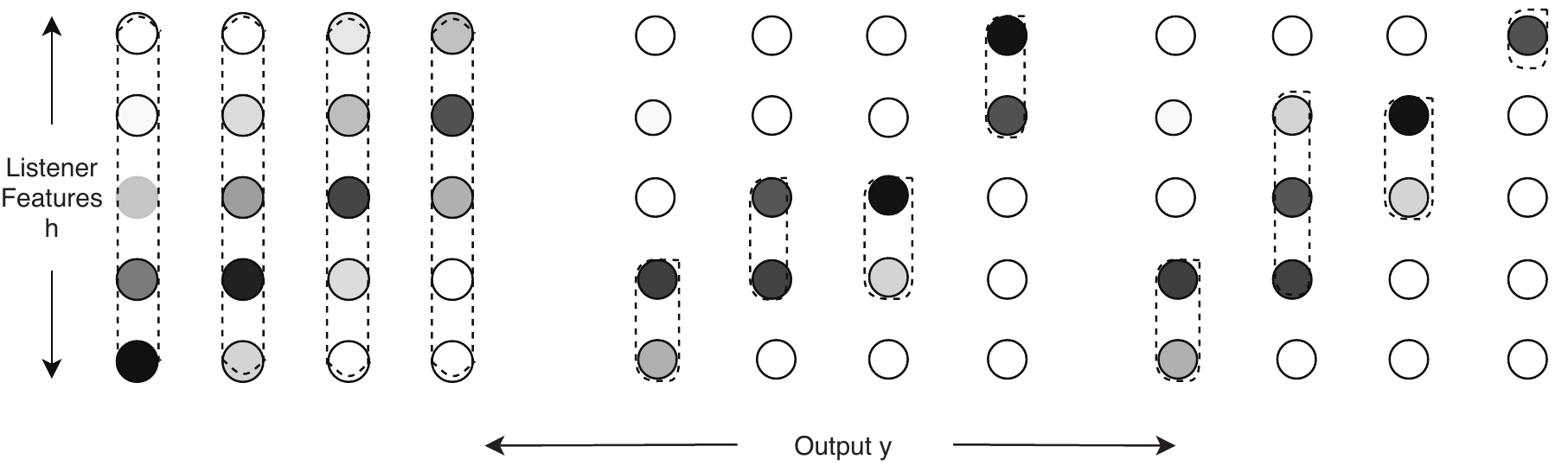}}
\centerline{\\\\\ (a) GSA \\\\\\\\ (b) MoChA \\\\\\ (c) AMoChA}
\caption{Comparison of different attention mechanisms. Each node represents the possibility of a listener feature (vertical axis) attended at a given output timestep (horizontal axis). The dotted box contains the required listener features to calculate $c_i$ at each step $y_i$. GSA need entire listener features to get $c_i$. MoChA uses fixed chunk length (2 in the figure) while AMoChA can learn chunk length adaptively at each step to calculate $c_i$.}
\label{fig:1}
\end{figure}

\subsection{Adaptive Monotonic Chunkwise Attender}
\label{ssec:blstm-amocha}

Global soft attention (GSA) stated in Section \ref{sec:basiclas} needs entire $U$ listener outputs to calculate $c_i$ at each step, which delay the computation in speller. To address the issue, a revised MoChA named Adaptive Monotonic Chunkwise Attention (AMoChA) is proposed to stream the attender. The difference between GSA and AMoChA is the calculation of $c_i$, which means Eq.{(\ref{eq:energy}) - (\ref{eq:context})} can be replaced with following process for inference and training respectively:
\subsubsection{Inference}
\label{sssec:inference}
For inference, we first compute an attend probability $p_{i,u}$ with listener output $h_u$ and speller state $s_i$ to represent the possibility of $h_u$ to be attended at step $i$. Then the attend/don't attend decision can be made by sampling from a Bernoulli random variable parameterized by $p_{i,u}$. To simplify the sampling process, we assume that $h_u$ should be attended if $p_{i,u}>0.5$ or $e_{i,u}>0$. After we get the attended $h_u$, which can be viewed as the boundary of listener outputs for each output step in speller, AMoChA computes the $c_i$ within an adaptive self-learned chunk ($W$) of listener outputs which starts from $h_{u-W+1}$ to $h_u$.
The above process can be formulated as:
\begin{equation}
e_{i,u} = Energy(s_i,h_u) = g\frac{v^T}{||v||}tanh(W_ss_i+W_hh_u+b)+r
\label{eq:modified_energy}
\end{equation}
\begin{gather}
\label{eq:sigmoid}
p_{i,u} = \sigma(e_{i,u}) \\
z_{i,u} \sim Bernoulli(p_{i,u}) \\
\label{eq:subscript}
v = u - W + 1 \\
d_{i,k} = Energy(s_i, h_k) \qquad k=v, v+1, ..., u \\
c_i = \sum_{k=v}^u \frac{exp(d_{i,k})}{\sum_{l=v}^u exp(d_{i,l})} h_k
\label{eq:mocha_soft}
\end{gather}
where $g$, $r$ are learnable scalars. The two energy functions above are similar but with different parameters. 

The difference between MoChA and AMoChA is the acquisition of chunk length $W$. For MoChA, fixed-length chunk $W$ is used and it is hard to be regulated. This hyper-parameter varies from language to output unit and is also affected by the pronunciation rate. To solve the problem, AMoChA is proposed to get free from selecting of hyper-parameter $W$.
These two mechanisms are visualized in Fig. \ref{fig:1}b, \ref{fig:1}c.

For AMoChA, after we get the attended $h_u$, two approaches are proposed to learn the chunk length $W$ with $h_u$ and speller state $s_i$ at step $i$. The ideas are similar with \cite{tjandra2017local}. But they use only $h_u$ as input and the usage of $W$ is different from us. 

\textbf{Constrained chunk length prediction}: The chunk length $W$ is bounded by the maximum of chunk length $W_{max}$ with the following equation:
\begin{equation}
W=W_{max}*\sigma(V_p^T F(W_h h_u+W_s s_i+b))
\label{eq:constrained}
\end{equation}
where, $W_h$, $W_s$, $V_p^T$ are parameters to learn, $\sigma$ is sigmoid function and activate function (F) can be ReLU \cite{glorot2011deep} or Tanh.

\textbf{Unconstrained chunk length prediction}: Compared to the Eq.(\ref{eq:constrained}), we investigate an unconstrained chunk length prediction network which ignores the hyper-parameter $W_{max}$:
\begin{equation}
W=exp(V_p^T F(W_h h_u+W_s s_i+b))
\label{eq:unconstrained}
\end{equation}
where the exponential function here can ensure the chunk length $W$ to be positive.

In the original paper of MoChA, the author also talked about its limitation of fixed-length chunk and proposed Monotonic Adaptive Chunkwise Attention (MATCHA). The method implements adaptive chunk length by using the chunk within two adjacent attended listener outputs and calculate expectation of $c_i$ as MoChA for training. On the contrary, our method use several neural layers to predict the chunk length directly.

\subsubsection{Training}
\label{sssec:training}
For MoChA training, the model can not be trained with back-propagation due to the sampling process. To remedy this, Raffel et al. propose to get $c_i=\sum_u \beta_{i,u}h_u $ by computing the probability of weight distribution over listener outputs, which can be formulated as:
\begin{gather}
\label{eq:expectation a}
\alpha_{i,u} = p_{i,u}((1-p_{i,u-1}) \frac{\alpha_{i, u-1}}{p_{i, u-1}}+\alpha_{i-1,u}) \\
\label{eq:expectation b}
\beta_{i,u} = \sum_{k=u}^{u+W-1}\frac{\alpha_{i,k}exp(d_{i,u})}{\sum_{l=k-W+1}^k{exp(d_{i,l})}}
\end{gather}
where the computation of $\alpha_{i,u}$ and $\beta_{i,u}$ are time costed. 
There is an efficient parallelized algorithm stated in \cite{chiu2017monotonic}. We are not going to discuss it here due to the page limitation. 

In order to learn parameters in prediction network of AMoChA, we use a multi-task loss function that is composed of standard cross-entropy (CE) loss and mean square error (MSE) loss between the predicted chunk length and the ground truth chunk length, which is as follows:
\begin{equation}
\label{eq:mse}
Loss=(1 - \lambda) \times L_{CE}+\lambda \times L_W
\end{equation}
where $\lambda$ controls  the ratio of  CE and MSE.
The ground truth chunk length for each output character can be obtained from greedy decoding or beam search decoding of each utterance with BLSTM-GSA. We use the number of attention weight that is above a threshold (0.01) as the target chunk length. Character level alignment from the hybrid system can also be used to obtain the true chunk length for each character.


\section{EXPERIMENTS}
\label{sec:results}
Our experiments are conducted on a $\sim$ 1,000 hour Sogou internal data from voice dictation. The results are average WER of three test sets which have about 22,000 utterances in total.

All experiments use 40-dimensional filter bank features, computed every 10ms within a 25ms window. First and second derivates are not used. Similar to \cite{sak2015fast}, we concatenate every 5 consecutive frames to form a 200-dimensional feature vector as the input to our network. 

For our LAS baseline, a four layer BLSTM with 256 hidden units on each LSTM is used as encoder. The third and forth layer of BLSTM uses a pyramid structure that takes every two consecutive frames of its input as input. As a result, the final output representation of listener is 4x subsampled. Addictive attention \cite{bahdanau2014neural} is used in the attender. The speller is a 1-layer LSTM with 512 hidden units. To compare the performance of LC listener, we also explore the LSTM-GSA model with unidirectional LSTMs encoder \cite{hochreiter1997long} whose hidden units is 640. The number of parameters is approximately the same as BLSTM-GSA. The hidden size of energy function and prediction network in AMoChA are both 512.

The total number of modelling unit is 6812, including 6809 Chinese characters, start of sequence (SOS), end of sequence (EOS) and unknown character (UNK). All models are trained by optimizing the CE loss except for the AMoChA which is learned by optimizing Eq(\ref{eq:mse}).
Scheduled sampling \cite{bengio2015scheduled} and label smoothing \cite{chorowski2016towards} are all adopted during training to improve performance. 
The training epoch for BLSTM-GSA and LSTM-GSA is 30.
We use teacher force at the first 11 epochs and
use output of last step to feed into network with schedule sampling
rate that gradually increases to 0.3 from epoch 12 to epoch 17. From
epoch 17, we fix schedule sampling rate to 0.3. We use an initial learning rate of 0.0002 and halve it from epoch 24. 
The other models are trained for 14 epochs with similar schedule sampling and learning rate setting. 
Beam search decoding is used without an external language model to evaluate our model on test sets and the beam width is set to 5. Temperature \cite{chorowski2016towards} is also used in decoding to let the output distribution more smooth for better beam search results. The weight decay is 1e-5 \cite{cortes20092}.

\subsection{LC-GSA: Streaming the Listener}
\label{ssec:results_lc}
We start to train BLSTM-GSA (MDL1) and LSTM-GSA (MDL2) as our baselines. The LSTM-GSA performs poor as seen in Table \ref{table:LC-GSA-LAS}. Then we explore the ability of LC-BLSTM to stream the listener. As stated in section \ref{ssec:lcblstm}, the input sequence is split into none-overlapping blocks with length $N_c$. Then $N_r$ future frames are appended to each block. 
The value of $N_r$ controls the latency of model. Table \ref{table:LC-GSA-LAS} shows the results of our methods and it indicates that training from BLSTM-GSA can result in an impressive improvement than training from scratch. If $N_c$ and $N_r$ are chosen properly (64 and 32 for our task), the LC-GSA (MDL3) behaves nearly the same as MDL1.

With LC-GSA, we think attention weight can be redistributed to attend more future frames and get similar context vector as BLSTM-GSA at each output step even if the LC-BLSTM lost some future information.
\begin{table}[th]
\caption{Results of our methods to stream the listener}
\label{table:LC-GSA-LAS}
\centering
\begin{tabular}{c c c c c}
\toprule 
\textbf{Model} & $\mathbf{N_c}$ & $\mathbf{N_r}$ & \textbf{Initial Model} & \textbf{CER(\%)} \\
\midrule
MDL1 & - & - & Null & $17.88$ \\
MDL2  & - & - & Null & $20.4$ \\
\midrule
MDL3 & $32$ & $16$ & Null & $20.13$\\
MDL3 & $32$ & $16$ & MDL1 & $18.63$\\
MDL3 & $64$ & $32$ & MDL1 & $\mathbf{17.99}$\\
\bottomrule
\end{tabular}
\end{table}

\subsection{BLSTM-AMoChA: Streaming the Attender}
\label{ssec:results_amocha}
Next, we investigate the effectiveness of our proposed AMoChA with an BLSTM listener. The first experiment is the implementation of MoChA with BLSTM listener, denoted as BLSTM-MoChA (MDL4). 
Our best model of MDL4 is trained with MDL1 as initial model and the setting of chunk length is 10 (40 frames of the raw input in listener) for Chinese characters. It achieves a CER of 17.7\% which is comparable to MDL1.

Table \ref{table:blstm-amocha} compares the results of two different chunk length prediction methods, which are proposed in section \ref{ssec:blstm-amocha}. These models are all trained with MDL1 as initial model. For constrained chunk length prediction method, we denote it as Constrained BLSTM-AMoChA (C-MDL5). The table shows that the best upper bound of chunk length is 40 and the 0.02 is the best proportion of the chunk length prediction task. For the activate function, ReLU behaves slightly better than Tanh. With the best setting, Unconstrained BLSTM-AMoChA (U-MDL5) with different ground truth chunk length are studied. BS5 seems provide the most accurate chunk length supervision. The best result of model U-MDL5 can achieve a 17.36\% of CER and outperform the MDL4, which use fixed chunk length.
\begin{table}[t]
\caption{The results of BLSTM-AMoChA. F is the activate function in Eq.\ref{eq:constrained}. SOL (source of label) illustrates how we get the ground truth chunk length. BS1 means greedy decoding with MDL1. BS5 represents beam decoding with beam size equals 5 and extra language model. HMM is the alignment result of hybrid system.}
\label{table:blstm-amocha}
\centering
\begin{tabular}{c c c c c c}
\toprule
\textbf{Model} & $\mathbf{W_{max}}$ & $\mathbf{\lambda}$ & \textbf{F} & \textbf{SOL} & \textbf{CER(\%)} \\
\midrule
MDL1 & - & $0$ & - & - & $17.88$ \\
MDL4 & - & $0$ & - & - & $17.77$ \\
\midrule
C-MDL5 & $30$ & $0.1$ & ReLU & BS1 & $17.8$ \\
C-MDL5 & $40$ & $0.1$ & ReLU & BS1 & $17.75$ \\
C-MDL5 & $60$ & $0.1$ & ReLU & BS1  & $17.92$ \\
C-MDL5 & $40$ & $0.2$ & ReLU & BS1  & $17.9$ \\
C-MDL5 & $40$ & $0.02$ & ReLU & BS1  & $17.51$ \\
C-MDL5 & $40$ & $0.02$ & Tanh & BS1 & $17.54$ \\
\midrule
U-MDL5 & - & $0.02$ & ReLU & BS1 & $17.45$ \\
U-MDL5 & - & $0.02$ & ReLU & HMM & $17.40$ \\
U-MDL5 & - & $0.02$ & ReLU & BS5 & $\mathbf{17.36}$ \\
\bottomrule
\end{tabular}
\end{table}

\subsection{LC-AMoChA: Online LAS}
\label{ssec:results_online}
After stream the listener and attender separately, we then combine the two parts with best setting and acquire LC-MoChA (MDL6) and LC-AMoChA (MDL7). These two models behave not well if no extra methods are used. In this part, we will figure out why it will happen and explore how to solve the problem. 

\begin{figure}[tbp]
\centering
\begin{minipage}[b]{0.32\linewidth}
\centerline{\includegraphics[width=2.8cm]{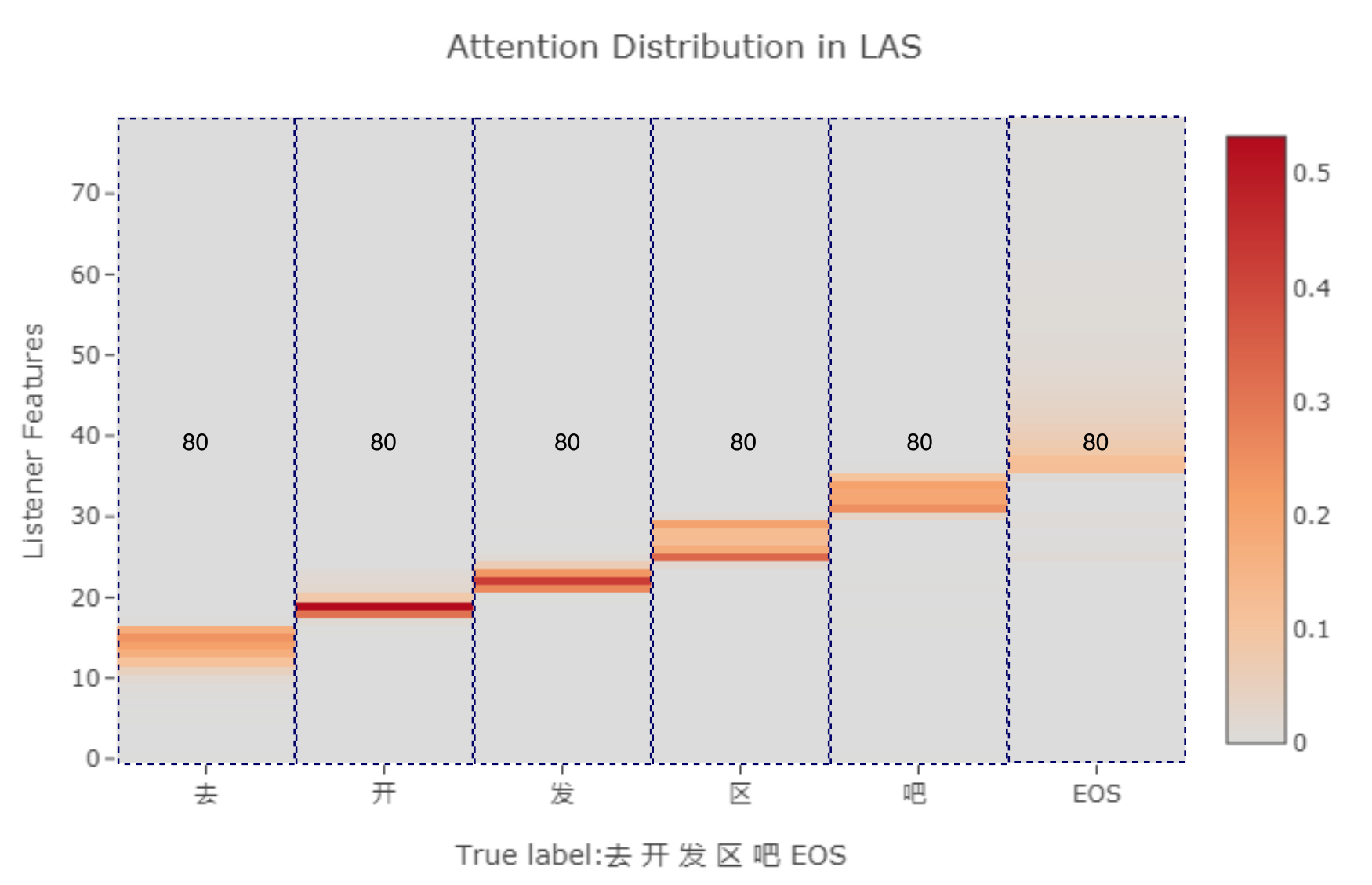}}
\centerline{(a) MDL1}\medskip
\end{minipage}
\hfill
\begin{minipage}[b]{0.32\linewidth}
\centering
\centerline{\includegraphics[width=2.8cm]{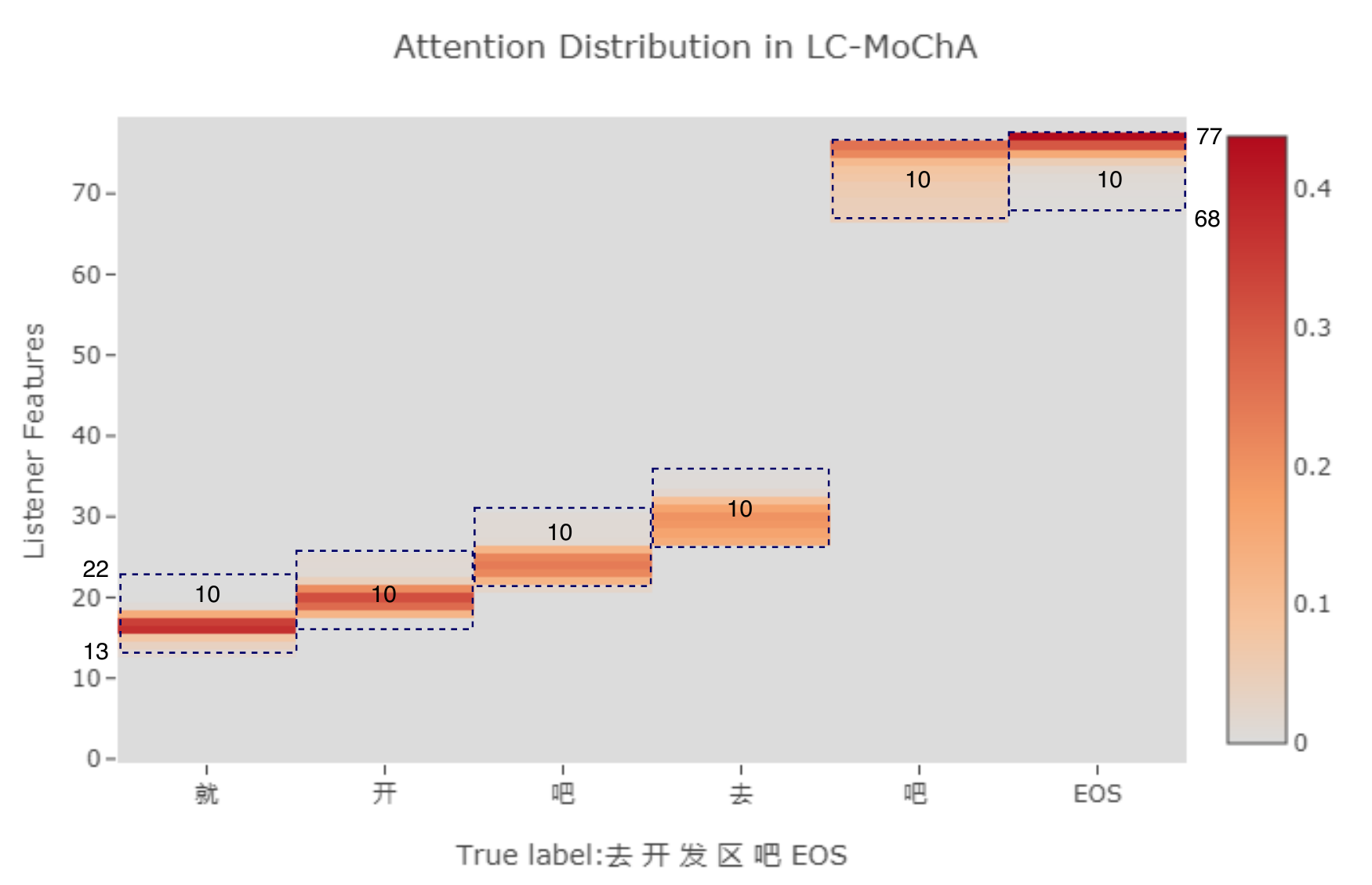}}
\centerline{(b) MDL6}\medskip
\end{minipage}
\hfill
\begin{minipage}[b]{0.32\linewidth}
\centering
\centerline{\includegraphics[width=2.8cm]{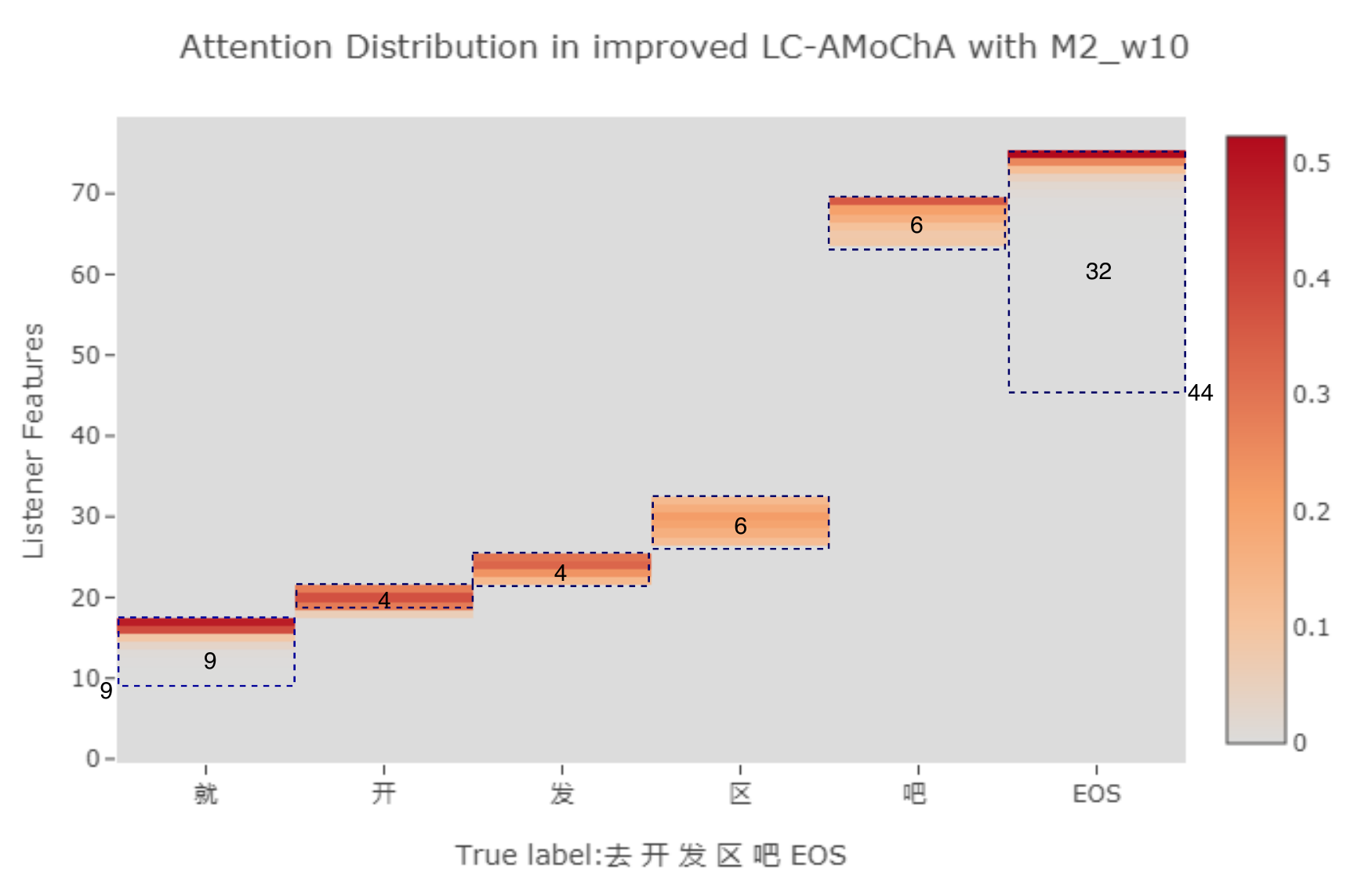}}
\centerline{(c) MDL7}\medskip
\end{minipage}
\caption{Distribution of attention weight in different models. The dotted box contains listener features with non-zero weight and chunk length is printed in the box. MDL6 is the model with only pre-training methods. MDL7 is the model in Table \ref{table:onlinelas}.}
\label{fig:2}
\end{figure}

According to the experience in section \ref{ssec:results_lc}, we use pre-trained model as initial model for MDL6 and MDL7. The results are exhibited in Table \ref{table:onlinelas}. Despite using pre-trained models, a huge degradation occurs (-13.1\%). By observing the attention weight distribution of MDL1 and MDL6, we find that the boundary of alignment shifts to the future listener features obviously. One example shows in Fig.\ref{fig:2}a, \ref{fig:2}b. MoChA calculates an energy scalar $p_{i,u}$ with listener features to determine the boundary of each output units. The lost future information of listener features caused by LC-BLSTM will make the boundary shift to the future and eventually lead to the degradation of CER. In order to compensate the degradation, we propose two methods to ease the problem.

\begin{table}[tbp]
\caption{Improvements to entire LAS. Future w means w-1 of future information are averaged.}
\label{table:onlinelas}
\centering
\begin{tabular}{c c c l}
\toprule
\textbf{Model} & \textbf{Methods} & \textbf{Future $\mathbf{w}$} & \textbf{CER(\%)}\\
\midrule
MDL1 & - & - & $17.88$ \\
MDL6 & - & 1 & $20.22(-13.1\%)$\\
\midrule
MDL6 & M1 & $8$ & $19.49$ \\
MDL6 & M1 & $10$ & $19.94$\\
MDL6 & M2 & $8$ & $18.87$\\
MDL6 & M2 & $10$ & $\mathbf{18.78}(-5\%)$\\
\midrule

MDL7 & M2 & $10$ & $\mathbf{18.5}(-3.5\%)$\\
\bottomrule
\end{tabular}
\end{table}

Given the listener outputs $\boldsymbol{h}$, we suppose the attended probability of $\boldsymbol{h}$ at step $i$ is $\boldsymbol{p_i}=\{p_{i,1},p_{i,2},\dots,p_{i,U}\}$, the first method (M1) we used is to append $w-1$ future listener features for each $h_u$ with the following equation to calculate $\boldsymbol{p_i}$:
\begin{equation}
\label{eq:M1}
\hat{h_u} = \frac{h_u + h_{u+1} + \dots + h_{u+w-1}}{w} \quad for     \quad u=1,2,\dots,U
\end{equation}

The second method (M2) is to utilize the future information for each attended probability $p_{i,u}$ at output step i:
\begin{equation}
\label{eq:M2}
\hat{p_{i,u}} = \frac{p_{i,u} + p_{i,u+1} + \dots + p_{i,u+w-1}}{w} \quad for \quad u=1,2,\dots,U
\end{equation}
It means the average probability is used to judge the boundary instead of the original $p_{i,u}$. The Eq.(\ref{eq:M2}) can be inserted after Eq.(\ref{eq:sigmoid}) during both training and inference.

By utilizing future information, we can drag back the boundary of alignment to some extent and the attention weights can be redistributed within the chunk length to have a better performance, which can be seen in Fig.\ref{fig:2}c. Table \ref{table:onlinelas} compares the result of our two revisions to MDL6 with only pre-training method. The table indicates that M2 with averaging 10 consecutive future attended possibilities can stream the MDL1 with least degradation (-5\%). Finally, together with AMoChA, the basic offline LAS can decode in a streaming manner with an acceptable 3.5\% relative degradation of CER.

\section{CONCLUSIONS}
\label{sec:conclusion}
In this paper, we explore an online attention-based LAS system. On the listener side, we propose to use LC-BLSTM to decrease the latency. On the attention side, we propose an AMoChA method to stream the attention and better fit the speech properties. By combining the two parts and utilizing our proposed methods, the BLSTM-GSA can be online with only a 3.5\% relative degradation of performance on our Mandarin corpus. In AMoChA, the chunk length at each step may be related to the last step. Perhaps LSTMs are better to be used here, which is included in our future work. Furthermore, we use an one-layer LSTM speller in our system. How to use a multi-layer speller with AMoChA is another challenge for online ASR systems.

\bibliographystyle{IEEEtran}
\bibliography{mybib}

\begin{thebibliography}{10}
\providecommand{\url}[1]{#1}
\csname url@samestyle\endcsname
\providecommand{\newblock}{\relax}
\providecommand{\bibinfo}[2]{#2}
\providecommand{\BIBentrySTDinterwordspacing}{\spaceskip=0pt\relax}
\providecommand{\BIBentryALTinterwordstretchfactor}{4}
\providecommand{\BIBentryALTinterwordspacing}{\spaceskip=\fontdimen2\font plus
\BIBentryALTinterwordstretchfactor\fontdimen3\font minus
  \fontdimen4\font\relax}
\providecommand{\BIBforeignlanguage}[2]{{%
\expandafter\ifx\csname l@#1\endcsname\relax
\typeout{** WARNING: IEEEtran.bst: No hyphenation pattern has been}%
\typeout{** loaded for the language `#1'. Using the pattern for}%
\typeout{** the default language instead.}%
\else
\language=\csname l@#1\endcsname
\fi
#2}}
\providecommand{\BIBdecl}{\relax}
\BIBdecl

\bibitem{graves2006connectionist}
A.~Graves, S.~Fern{\'a}ndez, F.~Gomez, and J.~Schmidhuber, ``Connectionist
  temporal classification: labelling unsegmented sequence data with recurrent
  neural networks,'' in \emph{Proceedings of the 23rd international conference
  on Machine learning}.\hskip 1em plus 0.5em minus 0.4em\relax ACM, 2006, pp.
  369--376.

\bibitem{amodei2016deep}
D.~Amodei, S.~Ananthanarayanan, R.~Anubhai, J.~Bai, E.~Battenberg, C.~Case,
  J.~Casper, B.~Catanzaro, Q.~Cheng, G.~Chen \emph{et~al.}, ``Deep speech 2:
  End-to-end speech recognition in english and mandarin,'' in
  \emph{International conference on machine learning}, 2016, pp. 173--182.

\bibitem{graves2012sequence}
A.~Graves, ``Sequence transduction with recurrent neural networks,''
  \emph{arXiv preprint arXiv:1211.3711}, 2012.

\bibitem{battenberg2017exploring}
E.~Battenberg, J.~Chen, R.~Child, A.~Coates, Y.~G.~Y. Li, H.~Liu, S.~Satheesh,
  A.~Sriram, and Z.~Zhu, ``Exploring neural transducers for end-to-end speech
  recognition,'' in \emph{2017 IEEE Automatic Speech Recognition and
  Understanding Workshop (ASRU)}.\hskip 1em plus 0.5em minus 0.4em\relax IEEE,
  2017, pp. 206--213.

\bibitem{rao2017exploring}
K.~Rao, H.~Sak, and R.~Prabhavalkar, ``Exploring architectures, data and units
  for streaming end-to-end speech recognition with rnn-transducer,'' in
  \emph{2017 IEEE Automatic Speech Recognition and Understanding Workshop
  (ASRU)}.\hskip 1em plus 0.5em minus 0.4em\relax IEEE, 2017, pp. 193--199.

\bibitem{sak2017recurrent}
H.~Sak, M.~Shannon, K.~Rao, and F.~Beaufays, ``Recurrent neural aligner: An
  encoder-decoder neural network model for sequence to sequence mapping.'' in
  \emph{Interspeech}, 2017, pp. 1298--1302.

\bibitem{dong2018extending}
L.~Dong, S.~Zhou, W.~Chen, and B.~Xu, ``Extending recurrent neural aligner for
  streaming end-to-end speech recognition in mandarin,'' \emph{arXiv preprint
  arXiv:1806.06342}, 2018.

\bibitem{yu2016online}
L.~Yu, J.~Buys, and P.~Blunsom, ``Online segment to segment neural
  transduction,'' \emph{arXiv preprint arXiv:1609.08194}, 2016.

\bibitem{chorowski2014end}
J.~Chorowski, D.~Bahdanau, K.~Cho, and Y.~Bengio, ``End-to-end continuous
  speech recognition using attention-based recurrent nn: First results,''
  \emph{arXiv preprint arXiv:1412.1602}, 2014.

\bibitem{chan2016listen}
W.~Chan, N.~Jaitly, Q.~Le, and O.~Vinyals, ``Listen, attend and spell: A neural
  network for large vocabulary conversational speech recognition,'' in
  \emph{2016 IEEE International Conference on Acoustics, Speech and Signal
  Processing (ICASSP)}.\hskip 1em plus 0.5em minus 0.4em\relax IEEE, 2016, pp.
  4960--4964.

\bibitem{chiu2018state}
C.-C. Chiu, T.~N. Sainath, Y.~Wu, R.~Prabhavalkar, P.~Nguyen, Z.~Chen,
  A.~Kannan, R.~J. Weiss, K.~Rao, E.~Gonina \emph{et~al.}, ``State-of-the-art
  speech recognition with sequence-to-sequence models,'' in \emph{2018 IEEE
  International Conference on Acoustics, Speech and Signal Processing
  (ICASSP)}.\hskip 1em plus 0.5em minus 0.4em\relax IEEE, 2018, pp. 4774--4778.

\bibitem{schuster1997bidirectional}
M.~Schuster and K.~K. Paliwal, ``Bidirectional recurrent neural networks,''
  \emph{IEEE Transactions on Signal Processing}, vol.~45, no.~11, pp.
  2673--2681, 1997.

\bibitem{prabhavalkar2017comparison}
R.~Prabhavalkar, K.~Rao, T.~N. Sainath, B.~Li, L.~Johnson, and N.~Jaitly, ``A
  comparison of sequence-to-sequence models for speech recognition.'' in
  \emph{Interspeech}, 2017, pp. 939--943.

\bibitem{zhang2016highway}
Y.~Zhang, G.~Chen, D.~Yu, K.~Yaco, S.~Khudanpur, and J.~Glass, ``Highway long
  short-term memory rnns for distant speech recognition,'' in \emph{2016 IEEE
  International Conference on Acoustics, Speech and Signal Processing
  (ICASSP)}.\hskip 1em plus 0.5em minus 0.4em\relax IEEE, 2016, pp. 5755--5759.

\bibitem{xue2017improving}
S.~Xue and Z.~Yan, ``Improving latency-controlled blstm acoustic models for
  online speech recognition,'' in \emph{2017 IEEE International Conference on
  Acoustics, Speech and Signal Processing (ICASSP)}.\hskip 1em plus 0.5em minus
  0.4em\relax IEEE, 2017, pp. 5340--5344.

\bibitem{jaitly2016online}
N.~Jaitly, Q.~V. Le, O.~Vinyals, I.~Sutskever, D.~Sussillo, and S.~Bengio, ``An
  online sequence-to-sequence model using partial conditioning,'' in
  \emph{Advances in Neural Information Processing Systems}, 2016, pp.
  5067--5075.

\bibitem{sainath2018improving}
T.~N. Sainath, C.-C. Chiu, R.~Prabhavalkar, A.~Kannan, Y.~Wu, P.~Nguyen, and
  Z.~Chen, ``Improving the performance of online neural transducer models,'' in
  \emph{2018 IEEE International Conference on Acoustics, Speech and Signal
  Processing (ICASSP)}.\hskip 1em plus 0.5em minus 0.4em\relax IEEE, 2018, pp.
  5864--5868.

\bibitem{luo2017learning}
Y.~Luo, C.-C. Chiu, N.~Jaitly, and I.~Sutskever, ``Learning online alignments
  with continuous rewards policy gradient,'' in \emph{2017 IEEE International
  Conference on Acoustics, Speech and Signal Processing (ICASSP)}.\hskip 1em
  plus 0.5em minus 0.4em\relax IEEE, 2017, pp. 2801--2805.

\bibitem{lawson2018learning}
D.~Lawson, C.-C. Chiu, G.~Tucker, C.~Raffel, K.~Swersky, and N.~Jaitly,
  ``Learning hard alignments with variational inference,'' in \emph{2018 IEEE
  International Conference on Acoustics, Speech and Signal Processing
  (ICASSP)}.\hskip 1em plus 0.5em minus 0.4em\relax IEEE, 2018, pp. 5799--5803.

\bibitem{raffel2017online}
C.~Raffel, M.-T. Luong, P.~J. Liu, R.~J. Weiss, and D.~Eck, ``Online and
  linear-time attention by enforcing monotonic alignments,'' in
  \emph{Proceedings of the 34th International Conference on Machine
  Learning-Volume 70}.\hskip 1em plus 0.5em minus 0.4em\relax JMLR. org, 2017,
  pp. 2837--2846.

\bibitem{chiu2017monotonic}
C.-C. Chiu and C.~Raffel, ``Monotonic chunkwise attention,'' \emph{arXiv
  preprint arXiv:1712.05382}, 2017.

\bibitem{tjandra2017local}
A.~Tjandra, S.~Sakti, and S.~Nakamura, ``Local monotonic attention mechanism
  for end-to-end speech and language processing,'' \emph{arXiv preprint
  arXiv:1705.08091}, 2017.

\bibitem{glorot2011deep}
X.~Glorot, A.~Bordes, and Y.~Bengio, ``Deep sparse rectifier neural networks,''
  in \emph{Proceedings of the fourteenth international conference on artificial
  intelligence and statistics}, 2011, pp. 315--323.

\bibitem{sak2015fast}
H.~Sak, A.~Senior, K.~Rao, and F.~Beaufays, ``Fast and accurate recurrent
  neural network acoustic models for speech recognition,'' \emph{arXiv preprint
  arXiv:1507.06947}, 2015.

\bibitem{bahdanau2014neural}
D.~Bahdanau, K.~Cho, and Y.~Bengio, ``Neural machine translation by jointly
  learning to align and translate,'' \emph{arXiv preprint arXiv:1409.0473},
  2014.

\bibitem{hochreiter1997long}
S.~Hochreiter and J.~Schmidhuber, ``Long short-term memory,'' \emph{Neural
  computation}, vol.~9, no.~8, pp. 1735--1780, 1997.

\bibitem{bengio2015scheduled}
S.~Bengio, O.~Vinyals, N.~Jaitly, and N.~Shazeer, ``Scheduled sampling for
  sequence prediction with recurrent neural networks,'' in \emph{Advances in
  Neural Information Processing Systems}, 2015, pp. 1171--1179.

\bibitem{chorowski2016towards}
J.~Chorowski and N.~Jaitly, ``Towards better decoding and language model
  integration in sequence to sequence models,'' \emph{arXiv preprint
  arXiv:1612.02695}, 2016.

\bibitem{cortes20092}
C.~Cortes, M.~Mohri, and A.~Rostamizadeh, ``L 2 regularization for learning
  kernels,'' in \emph{Proceedings of the Twenty-Fifth Conference on Uncertainty
  in Artificial Intelligence}.\hskip 1em plus 0.5em minus 0.4em\relax AUAI
  Press, 2009, pp. 109--116.

\end{thebibliography}
\end{document}